\documentclass{article}
\usepackage{spconf,amsmath,graphicx}
\usepackage{amssymb}
\usepackage{xspace}
\usepackage{color}
\usepackage{multirow}
\usepackage{graphics}
\usepackage{adjustbox}
\usepackage{subfigure}
\usepackage{amsmath}
\usepackage{paralist} 
\usepackage{enumitem} 
\usepackage{booktabs} 
\usepackage{array}

\graphicspath{{figures/}}

\usepackage[pagebackref=true,breaklinks=true,letterpaper=true,colorlinks,citecolor=blue,linkcolor=blue,bookmarks=false]{hyperref}

\long\def\ignorethis#1{}

\usepackage{tabularx}

\definecolor{gray}{rgb}{0.5,0.5,0.5}
\definecolor{MyBlue}{rgb}{0,0,1.0}
\definecolor{MyYellow}{rgb}{0.9,0.9,0}
\definecolor{MyRed}{rgb}{0.8,0.2,0}
\definecolor{MyGreen}{rgb}{0,0.5,0.0}
\definecolor{MyGray}{rgb}{0.4,0.4,0.4}

\newlength\paramargin
\newlength\figmargin
\newlength\secmargin

\newcolumntype{L}[1]{>{\raggedright\let\newline\\\arraybackslash\hspace{0pt}}m{#1}}
\newcolumntype{C}[1]{>{\centering\let\newline\\\arraybackslash\hspace{0pt}}m{#1}}
\newcolumntype{R}[1]{>{\raggedleft\let\newline\\\arraybackslash\hspace{0pt}}m{#1}}

\usepackage{url}

\newcommand{\para }[1]{\medskip \noindent {\bf #1}}

\title{Weakly-supervised Caricature Face Parsing through Domain Adaptation}
\name{Wenqing Chu$^{\star}$, Wei-Chih Hung$^{\dagger}$, Yi-Hsuan Tsai$^{\ddagger}$, Deng Cai$^{\star}$, Ming-Hsuan Yang$^{\dagger}$}
\address{$^{\star}$State Key Lab of CAD\&CG, Zhejiang University, China\\
	 $^{\dagger}$University of California at Merced, USA\\
	 $^{\ddagger}$NEC Laboratories America
	 }
\begin{document}
%
\maketitle

\begin{abstract}
A caricature is an artistic form of a person's picture in which certain striking characteristics are abstracted or exaggerated in order to create a humor or sarcasm effect.
For numerous caricature related applications such as attribute recognition and caricature editing, face parsing is an essential pre-processing step that provides a complete facial structure understanding.
However, current state-of-the-art face parsing methods require large amounts of labeled data on the pixel-level and such process for caricature is tedious and labor-intensive.
For real photos, there are numerous labeled datasets for face parsing.
Thus, we formulate caricature face parsing as a domain adaptation problem, where real photos play the role of the source domain, adapting to the target caricatures.
%
%
Specifically, we first leverage a spatial transformer based network to enable shape domain shifts. 
A feed-forward style transfer network is then utilized to capture texture-level domain gaps.
With these two steps, we synthesize face caricatures from real photos, and thus we can use parsing ground truths of the original photos to learn the parsing model.
Experimental results on the synthetic and real caricatures demonstrate the effectiveness of the proposed domain adaptation algorithm.
Code is available at: \url{https://github.com/ZJULearning/CariFaceParsing}.
\end{abstract}
\section{Introduction}
\label{sec:intro}
A caricature is a picture rendered through sketching, pencil strokes, or artistic styles, where some characteristics of its subject are abstracted or exaggerated.
Caricature parsing aims at predicting a class label (i.e., mouth, nose, to name a few) for each pixel in the given image.
It is one of the fundamental problems in numerous caricature related applications, such as attribute recognition~\cite{HuoBMVC2018WebCaricature} and caricature editing~\cite{lewiner2011interactive}. 

\begin{figure}[t]
\vspace{-3mm}
	\footnotesize
	\centering
	\renewcommand{\tabcolsep}{0.5pt} 
	\renewcommand{\arraystretch}{0.15} 
	\begin{center}

		\begin{tabular}{ccc}
			\includegraphics[width= 0.16\textwidth]{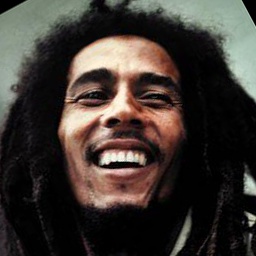}
			&	
			\includegraphics[width= 0.16\textwidth]{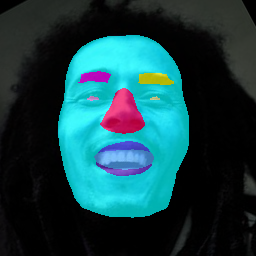}
			&	
			\includegraphics[width= 0.16\textwidth]{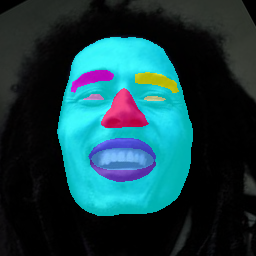}
			\\

			\includegraphics[width= 0.16\textwidth]{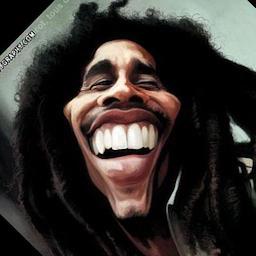}
			&	
			\includegraphics[width= 0.16\textwidth]{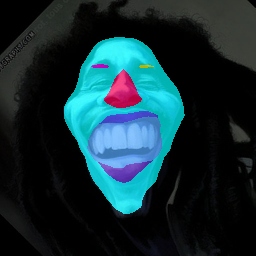}
			&	
			\includegraphics[width= 0.16\textwidth]{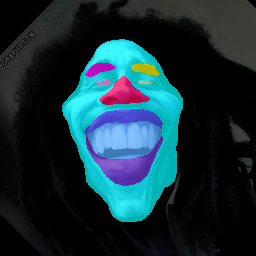}
			\\
			
			\addlinespace[0.15cm]
			Input Images &
			No Adaptation&
			Ours 
			\\
		\end{tabular}
	\end{center}
	\vspace{-0.2cm}		
	\caption{Example comparisons of our domain adaptation method. The first row is the real image and the second one is the real caricature. Although there exist large shape and texture differences between the photo and caricature, our model learned from the adapted source domain generalizes well on the real caricature.}
	\label{fig:motivation} 
	\vspace{-3mm}
\end{figure}

In recent years, deep face parsing models have received much attention and much progress has been made~\cite{luo2012hierarchical, liu2015multi, jackson2016cnn, Liu2017FacePV, wei2017learning}.
Luo et al.~\cite{luo2012hierarchical} develop a deep learning-based hierarchical model to parse a face from part/component levels to pixels.
An exemplar-based method is proposed in~\cite{smith2013exemplar}, where the aligned label maps of exemplary images are combined to obtain the final predictions.
In addition, semantic edge maps~\cite{liu2015multi} and landmark information~\cite{jackson2016cnn} are utilized as additional inputs to guide semantic part segmentation.
These methods rely on large-scale pixel-level face parsing ground truths to achieve good parsing performance.
However, the labeling process for caricature face parsing is tedious and labor-intensive.
%
We thus formulate the caricature face parsing as a domain adaptation problem, where real photos with ground truths are viewed as the source domain.

\begin{figure*}[!ht]
	\footnotesize
	\centering
	\renewcommand{\tabcolsep}{1pt} 
	\renewcommand{\arraystretch}{1} 
	\begin{center}
		\includegraphics[width= 0.8\textwidth]{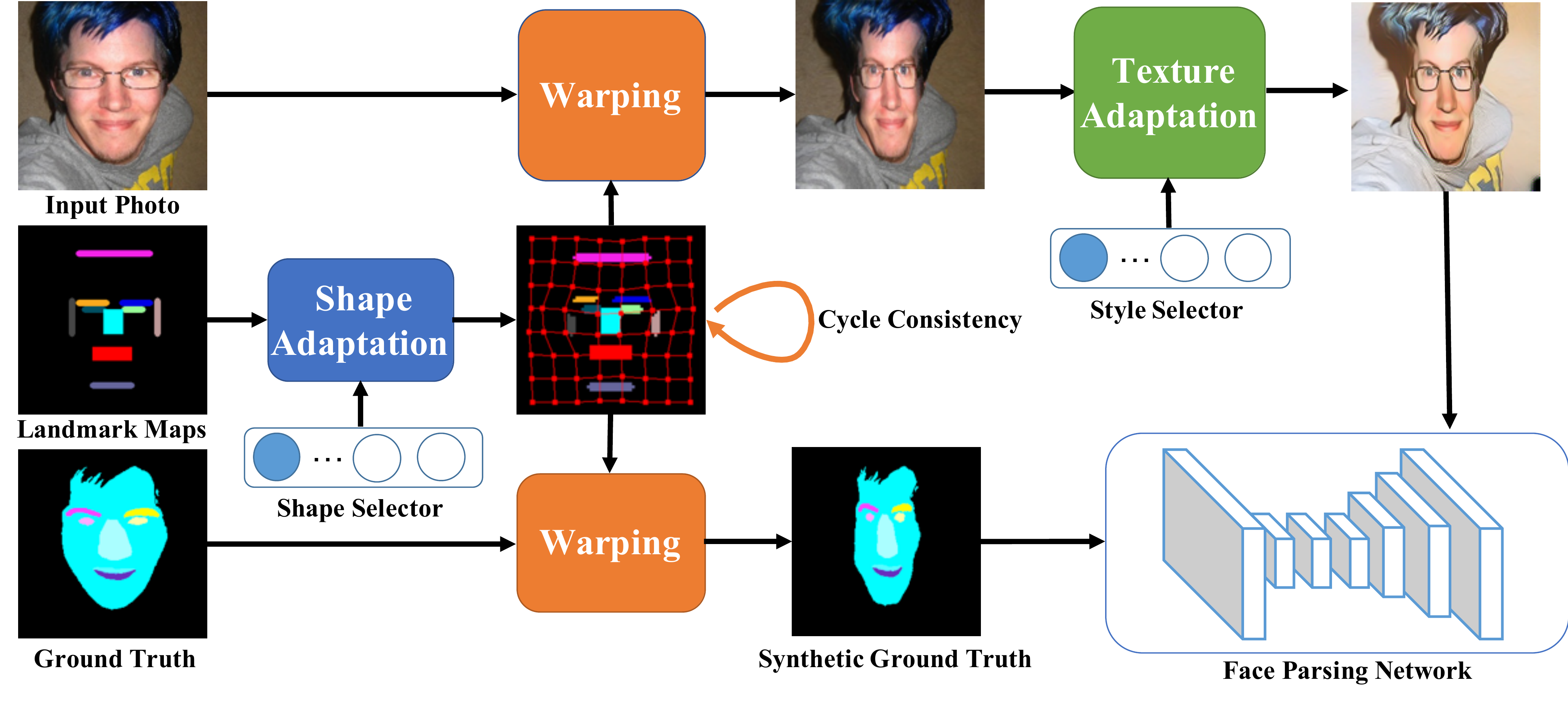}

	\end{center}
	\vspace{-0.4cm}
	\caption{Framework of the proposed domain adaptation algorithm. We propose to tackle the domain adaptation task by translating the face images into caricature training pairs with two modules: \textbf{shape adaptation} for synthesizing the shape exaggeration and \textbf{texture adaptation} for mimicking the style of caricature images.}
	\label{fig:framework} 
	\vspace{-0.2cm}
\end{figure*}

In this work, we propose to exploit available real-face parsing datasets to perform domain adaptation for learning a caricature face parsing model.
%
State-of-the-art domain adaptation approaches~\cite{hoffman2016fcns, chen2017no, hoffman2017cycada, tsai2018learning, pan2018two} utilize the adversarial network to learn domain invariant representations. 
Recently, CyCADA~\cite{hoffman2017cycada} employs image translation to adapt representations on the image level. 
This motivates us to perform image translation on facial photos to obtain caricature-like images, and then the face parsing model trained on the synthesized datasets would perform better on real-world caricatures. 
For image translation, however, state-of-the-art methods based on generative adversarial networks (GANs)~\cite{goodfellow2014generative} or neural style transfer~\cite{gatys2016image} can only capture pixel-level and low-level domain shifts between photos and caricatures.
%
%
The shape exaggeration in caricatures remains a great challenge for the visual domain adaptation. 
Therefore, we develop a spatial transformer based network to enable shape domain shifts, with the help from weak supervisions of caricature landmarks.
Then, a feed-forward style transfer network is utilized to capture texture-level domain gaps.
With these two steps, our framework is able to produce caricature-like images adapted from real face photos.


The contributions of this work are summarized as follows:
\begin{itemize}
	\setlength{\itemsep}{0pt}
	\setlength{\parsep}{0pt}
	\setlength{\parskip}{0.2pt}
	\item We present the first domain adaptation framework to learn the caricature face parsing model. 
	One example is shown in Fig.~\ref{fig:motivation}.
	\item The proposed framework adapts both shape and texture on the image level for generating synthesized caricature-like images.
	%
	\item We evaluate the proposed model on both synthetic and real caricature~\cite{HuoBMVC2018WebCaricature} datasets.
	Experimental results show that the proposed algorithm effectively generates high-quality parsing results compared to other domain adaptation methods. 
\end{itemize}
\vspace{-5mm}
\section{Proposed Method}
\vspace{-3mm}
\label{sec:pagestyle}
\begin{table*}[!t]
	\vspace{-3mm}
	\caption{Ablation study using different training datasets on the synthesized HELEN dataset.}
	\label{table:synthetic}
	\scriptsize
	\setlength{\belowcaptionskip}{-0.8cm}
	\newcommand{\tabincell}[2]{\begin{tabular}{@{}#1@{}}#2\end{tabular}}
	\newcolumntype{P}[1]{>{\centering\arraybackslash}p{#1}}
	
	\centering
	\begin{tabular}{m{2.2cm}  P{1.1cm} P{1.1cm} P{1.1cm} P{1.1cm} P{1.1cm} P{1.1cm}  P{1.1cm} P{1.1cm} P{1.1cm} P{1.1cm}}
		\toprule
		Methods & facial skin & eye-l & eye-r & brow-l & brow-r & nose &  in mouth &	upper lip &	lower lip  & avg \\
		\midrule
		HELEN  & 77.11 & 39.45 & 38.71 & 48.73 & 46.82 & 73.53 & 28.73 & 48.75 & 42.49 &  49.36\\
		HELEN-Cycle~\cite{zhu2017unpaired} & 81.55 & 47.60 & 46.85 & 59.44 & 59.19 & 79.47 & 37.62 & 56.12 & 51.29 & 57.68 \\
		HELEN-Texture  & 84.26 & 55.71 & 54.52 & 62.03 & 62.32 & 81.87 & 47.17 & 59.93 & 57.29 & 62.78 \\
		HELEN-Shape  & 85.68 & 53.28 & 54.17 & 64.52 & 64.99 & 83.50 & 52.15 & 65.88 & 62.63 & 65.19 \\
		HELEN-Cari (Ours)  & \textbf{89.01} & \textbf{63.94} & \textbf{64.42} & \textbf{70.58} & \textbf{72.98} & \textbf{87.67} & \textbf{64.33} & \textbf{74.46} & \textbf{73.04} & \textbf{73.38} \\
		\bottomrule
	\end{tabular} 
	
\end{table*}

\begin{table*}[!t]
	
	\caption{Comparison with state-of-the-art methods on the WebCaricature dataset~\cite{HuoBMVC2018WebCaricature}.}
	\label{table:caricature}
	\scriptsize
	\setlength{\belowcaptionskip}{-0.8cm}
	\newcommand{\tabincell}[2]{\begin{tabular}{@{}#1@{}}#2\end{tabular}}
	\newcolumntype{P}[1]{>{\centering\arraybackslash}p{#1}}
	
	\centering
	\begin{tabular}{m{2.2cm}  P{1.1cm} P{1.1cm} P{1.1cm} P{1.1cm} P{1.1cm} P{1.1cm}  P{1.1cm} P{1.1cm} P{1.1cm} P{1.1cm}}
		\toprule
		Methods & facial skin & eye-l & eye-r & brow-l & brow-r & nose &  in mouth &	upper lip &	lower lip  & avg \\
		\midrule
		HELEN  & 81.10 & 36.92 & 37.15 & 51.44 & 54.45 & 73.96 & 34.67 & 46.76 & 42.28 & 50.97 \\
		HELEN-Cycle~\cite{zhu2017unpaired} & 82.56 & 41.53 & 42.83 & 53.62 & 57.07 & 76.20 & 38.31 & 50.22 & 45.12 & 54.16 \\
		HELEN-Texture  & 83.67 & 45.92 & 47.68 & 55.32 & 58.46 & 78.20 & 44.97& 51.86 & 48.65 & 57.19 \\
		HELEN-Shape  & 85.31 & 48.45 & 50.99 & 58.48 & 60.74 & 79.90 & 48.09 & \textbf{63.42} & 53.42 & 60.97 \\
		Cari-Cross  & 81.81 & 45.91 & 44.00 & 60.80 & 61.21 & 77.65 & 36.04 & 48.44 & 50.76 &  56.29\\
		AdaptSegNet~\cite{tsai2018learning}  & 84.41 & 44.36 & 44.72 & 62.19 & 60.95 & 78.63 & 46.38 & 53.41 & 51.28 &  58.48\\
		HELEN-Cari (Ours)  & \textbf{86.54} & \textbf{54.93} & \textbf{56.73} & \textbf{63.67} & \textbf{65.10} & \textbf{82.07} & \textbf{55.55} & 58.63 & \textbf{54.92} & \textbf{64.23} \\
		\bottomrule
	\end{tabular} 
	
\end{table*}
In this section, we present a domain adaptation algorithm for caricature face parsing. 
We show the overall framework in Fig.~\ref{fig:framework}.
In this work, we perform domain adaptation on the image level, i.e., directly transform the source domain (face) training pairs to the target domain (caricature). 
We leverage image translation techniques to adapt the face photos to caricature-like images using two stages, including shape and texture adaptations. 
For shape adaptation, we propose a spatial transformation network to learn the shape exaggeration of caricature images.
For texture adaptation, we propose to train a conditional style transfer network to mimic the style of caricature images.
Since the synthesized caricatures are more similar to real caricatures in terms of shape and texture, the learned parsing model could generalize well on the target domain by training with the transformed training pairs.
%
%
\vspace{-3mm}
\subsection{Shape Adaptation}
\vspace{-2mm}
The proposed shape adaptation module is inspired by the Spatial Transformer Networks (STNs)~\cite{jaderberg2015spatial}. 
%
Different from conventional neural networks architectures, STNs reduce input geometric variations through an extra differentiable image warping operation whose parameters are predicted by a sub-network. 
In this work, our method also contains a differentiable image warping operation to perform plausible shape exaggeration on photos to capture shape domain shifts. 
Since there is no paired training data available, we train the shape adaptation module in a CycleGAN~\cite{zhu2017unpaired} fashion, where STNs are inserted into both the image translation networks.

In order to facilitate the training process, we do not use the diverse RGB images as input. 
Instead, our model learns to predict the warping parameters with one-hot landmark maps. 
We connect different ground truth landmarks into lines or regions such as mouth, nose, to name a few.
Then we treat each connected line or region as a single channel in the landmark map, which indicates the facial layout of the photo.
We show a colorized landmark map in Fig.~\ref{fig:framework}.
%
%
%
%

The shape adaptation network predicts the parameters for spatial warping on the photos to generate deformed ones.
We also perform the same transformation on the ground truth labels to obtain deformed ones correspondingly.
To better encode the diversity of shape exaggeration of caricatures, we leverage a one-hot vector as conditional input to generate diverse shapes. 
Before training, we first perform clustering on the caricature landmark positions and utilize the cluster centers as the shape sets.
Therefore, the shape adaptation network has two inputs, the landmark maps and a conditional input, where each bit in the conditional input represents a type in the shape sets. 
Note that, the weak supervision of landmark maps is only required during training, while at the test time, the model only needs the conditional input.

\vspace{-3mm}
\subsection{Texture Adaptation}
\vspace{-2mm}
After shape adaptation, the ensuing task is to equip the deformed photos with caricature appearances. 
Motivated by Li et al.~\cite{li2017diversified}, we also utilize conditional information to achieve diverse texture generation.
%
We first perform hierarchical clustering on the style features (detailed below) of all caricatures. Then we select those caricatures whose style features are close to cluster centers as reference caricature images.

The conditional style transfer network is based on the encoder-decoder architecture. 
We utilize a deformed photo and a randomly generated one-hot vector as the conditional input. 
To train the texture adaptation network, we combine the content loss and style loss similar to~\cite{li2017diversified}. 
We formulate the content loss as the feature differences between the generated caricature and the deformed photo at the $conv4\_2$ layer of the VGG model. 
The goal is to enforce the generated caricatures to preserve the structure of the deformed photos.
For the style loss, we compute the features by concatenating the statistics of Batch Normalization (BN)~\cite{li2017demystifying} instead of the Gram matrices \cite{simonyan2014very}.
%
Then we obtain the style loss as the feature differences between the generated caricature and the reference caricature image.
%
\vspace{-3mm}
\subsection{Caricature Face Parsing}
\vspace{-2mm}
After performing shape and texture adaptation to obtain synthesized caricature-like images and their transformed ground truth labels, we can train a caricature face parsing model.
We employ the ResNet-50~\cite{he2016deep} as the backbone.
%
%
To enlarge the receptive field, we apply dilations~\cite{yu2015multi} to the last two residual blocks and thus the output size of the backbone is $\frac{1}{8}$ of the input image size.
Furthermore, we include pyramid scene parsing module~\cite{zhao2017pyramid} to perform multi-scale feature ensembling. 
%

\vspace{-3mm}
\section{Experimental Results}
\label{sec:experimental}
\vspace{-3mm}

We first discuss the implementation details, experimental settings, datasets, and then present evaluation results.
%

\para{Implementation details.} 
Our framework and other baseline methods are implemented based on the PyTorch toolbox~\cite{paszke2017automatic}. 
For the shape adaptation and texture adaptation networks, we utilize ADAM~\cite{kingma2014adam} for training, where the learning rates are both set to $0.0001$. 
%
%
We use Wasserstein GAN~\cite{arjovsky2017wasserstein} to stabilize training for the shape adaptation network. 
%
For the image parsing network, we adopt the poly~\cite{chen2018deeplab, zhao2017pyramid} learning rate policy and set the initial learning rate as $0.001$.
%
 
\para{Datasets.} 
We employ the HELEN~\cite{smith2013exemplar, liu2015multi} dataset as the source domain. 
It contains $2330$ face photos with labeled facial components. 
The WebCaricature dataset~\cite{HuoBMVC2018WebCaricature} is used as the target domain. 
It contains $6042$ caricatures collected from the web and provides $17$ landmarks.
%
%
Since there are no face parsing labels in WebCaricature, we randomly select $100$ caricatures and manually annotate them for evaluation. 

\begin{figure*}[!ht]
	\footnotesize
	\centering
	\renewcommand{\tabcolsep}{1pt} 
	\renewcommand{\arraystretch}{1} 
	\begin{center}
		\begin{tabular}{cccccc}
			\includegraphics[width= 0.15\textwidth,height= 0.11\textwidth]{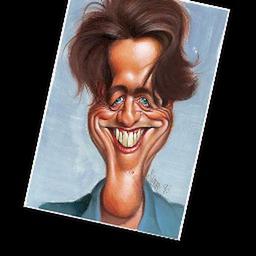}
			&	
			\includegraphics[width= 0.15\textwidth,height= 0.11\textwidth]{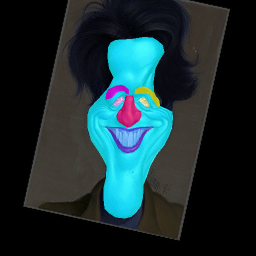}
			&	
			\includegraphics[width= 0.15\textwidth,height= 0.11\textwidth]{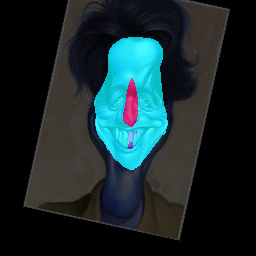}
			&
			\includegraphics[width= 0.15\textwidth,height= 0.11\textwidth]{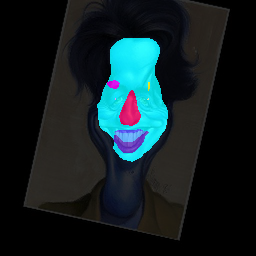}
			&
			\includegraphics[width= 0.15\textwidth,height= 0.11\textwidth]{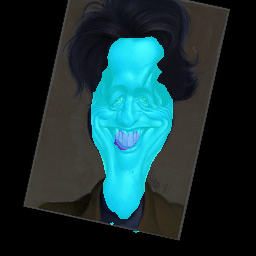}
			&	
			\includegraphics[width= 0.15\textwidth,height= 0.11\textwidth]{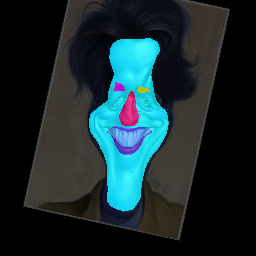}
			\\
			
			\includegraphics[width= 0.15\textwidth,height= 0.11\textwidth]{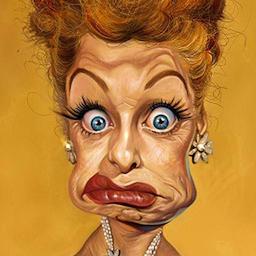}
			&	
			\includegraphics[width= 0.15\textwidth,height= 0.11\textwidth]{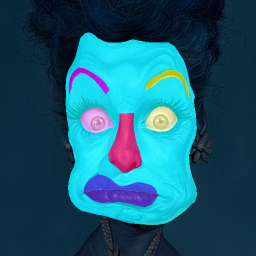}
			&	
			\includegraphics[width= 0.15\textwidth,height= 0.11\textwidth]{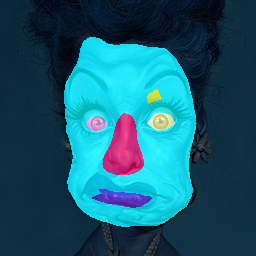}
			&
			\includegraphics[width= 0.15\textwidth,height= 0.11\textwidth]{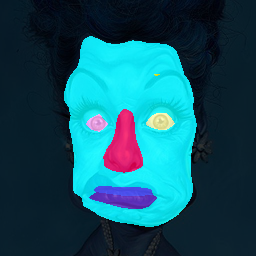}
			&
			\includegraphics[width= 0.15\textwidth,height= 0.11\textwidth]{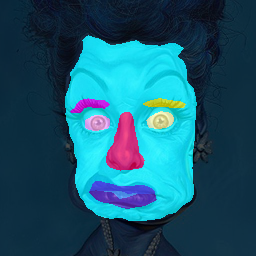}
			&	
			\includegraphics[width= 0.15\textwidth,height= 0.11\textwidth]{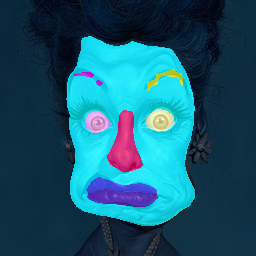}
			\\
			
			\includegraphics[width= 0.15\textwidth,height= 0.11\textwidth]{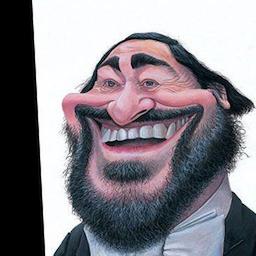}
			&	
			\includegraphics[width= 0.15\textwidth,height= 0.11\textwidth]{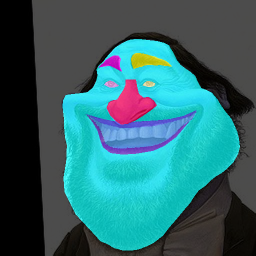}
			&	
			\includegraphics[width= 0.15\textwidth,height= 0.11\textwidth]{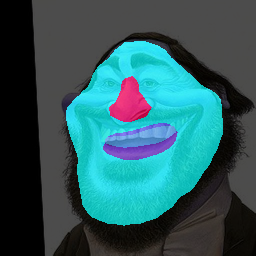}
			&
			\includegraphics[width= 0.15\textwidth,height= 0.11\textwidth]{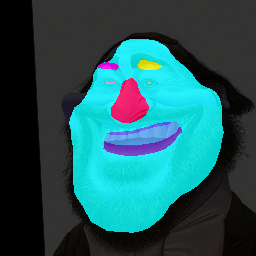}
			&
			\includegraphics[width= 0.15\textwidth,height= 0.11\textwidth]{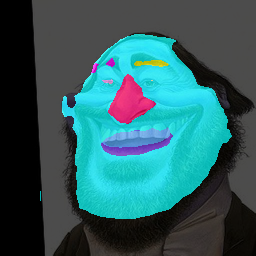}
			&	
			\includegraphics[width= 0.15\textwidth,height= 0.11\textwidth]{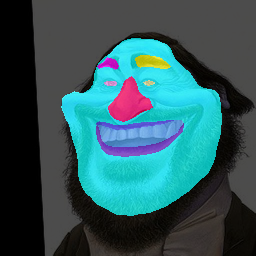}
			\\

			\includegraphics[width= 0.15\textwidth,height= 0.11\textwidth]{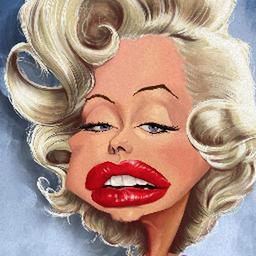}
			&	
			\includegraphics[width= 0.15\textwidth,height= 0.11\textwidth]{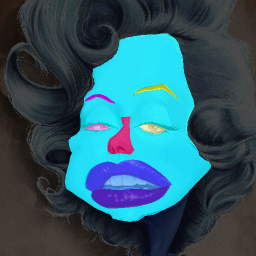}
			&	
			\includegraphics[width= 0.15\textwidth,height= 0.11\textwidth]{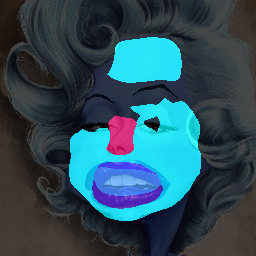}
			&
			\includegraphics[width= 0.15\textwidth,height= 0.11\textwidth]{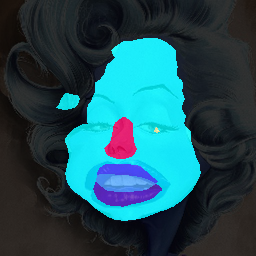}
			&
			\includegraphics[width= 0.15\textwidth,height= 0.11\textwidth]{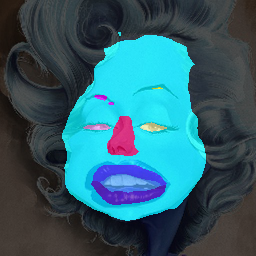}
			&	
			\includegraphics[width= 0.15\textwidth,height= 0.11\textwidth]{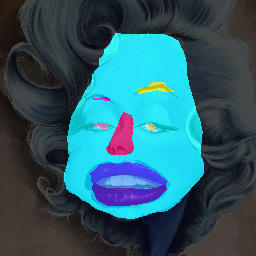}
			\\			

			\includegraphics[width= 0.15\textwidth,height= 0.11\textwidth]{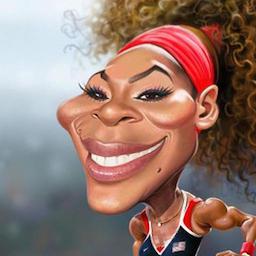}
			&	
			\includegraphics[width= 0.15\textwidth,height= 0.11\textwidth]{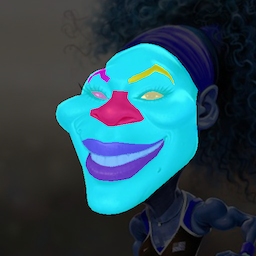}
			&	
			\includegraphics[width= 0.15\textwidth,height= 0.11\textwidth]{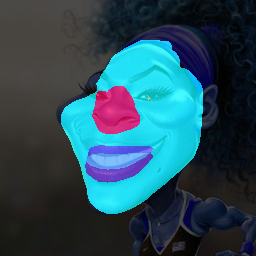}
			&
			\includegraphics[width= 0.15\textwidth,height= 0.11\textwidth]{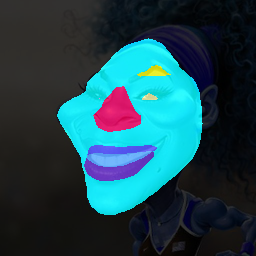}
			&
			\includegraphics[width= 0.15\textwidth,height= 0.11\textwidth]{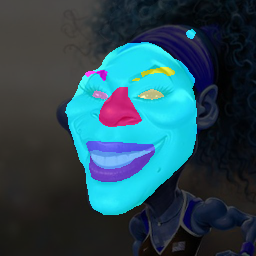}
			&	
			\includegraphics[width= 0.15\textwidth,height= 0.11\textwidth]{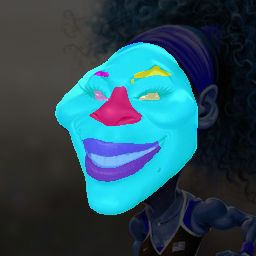}
			\\			
			
			\addlinespace[0.15cm]
			Input Images &
			Ground-Truth &
			No Adaptation &
			CycleGan~\cite{zhu2017unpaired} &
			AdaptSegNet~\cite{tsai2018learning} &
			Ours 
			\\
		\end{tabular}
	\end{center}
	\vspace{-0.4cm}
	\caption{Example results on the Webcaricature dataset~\cite{HuoBMVC2018WebCaricature}.}
	\label{fig:results} 
	\vspace{-0.2cm}
\end{figure*}

\vspace{-3mm}
\subsection{Ablation Study}
\vspace{-2mm}
In this section, we conduct an ablation study on a synthesized dataset denoted as HELEN-Cari obtained by performing the proposed transformation method on the entire HELEN dataset. 
%
%
%
Following the split in~\cite{smith2013exemplar, liu2015multi}, we use $330$ synthesized images in the HELEN-Cari dataset as the test set.
Then, we separately perform shape and texture adaptation on the training set of the HELEN dataset and denote them as HELEN-Texture and HELEN-Shape, respectively. 
To compare with CyCADA-like~\cite{hoffman2017cycada} methods, we apply CycleGan~\cite{zhu2017unpaired} to the training set of the HELEN dataset to generate the HELEN-Cycle dataset for comparison.

 In the following, we train the proposed image parsing networks with different training datasets and evaluate them on the HELEN-Cari dataset. 
 %
 %
 Table~\ref{table:synthetic} shows the ablation study with per-class IoU and mIoU for 9 classes as the metrics.
 The model trained only on the source HELEN dataset generalize poorly on the HELEN-Cari dataset,
 while the model trained with HELEN-Cari achieves the best performance, achieving $73.38$ mIoU. 
 The study shows that both the shape and texture adaptation contribute to the performance significantly. 
 The CycleGan~\cite{zhu2017unpaired} based domain adaptation method also helps achieve some performance gain.  
 However, it cannot handle the shape exaggeration in caricatures since it is only tailored for general image translation tasks.

\vspace{-3mm}
\subsection{Comparison with State-of-the-art Methods}
\vspace{-2mm}
We evaluate the proposed model on the manually-annotated caricatures on Webcaricature and present the results in Table~\ref{table:caricature}. 
Similarly, we show the evaluation results on models trained with the HELEN-Texture, HELEN-Shape, HELEN-Cycle and HELEN-Cari dataset. 
Moreover, we compare the proposed method with AdaptSegNet~\cite{tsai2018learning}, which is the state-of-the-art method for domain adaptation on semantic segmentation.
%
%
To further understand the need for real ground truths, we train an image parsing model denoted as Cari-Cross on the manually-annotated caricatures through cross-validation. 
%
The Cari-Cross model performs worse than most baseline methods with the mIoU as $56.29$ due to the insufficient training data.
With the proposed method, the model trained with the HELEN-Cari dataset performs favorably against other baseline methods with $64.23$ mIoU.
To qualitatively evaluate the results, we present visual comparisons on the realistic caricatures in Fig.~\ref{fig:results}. 
As shown in the figure, the CycleGan and AdaptSegNet can generate better parsing results than the model without adaptation. 
However, there are many failure cases around eyebrows and mouths due to the large shape and texture variations.
In contrast, the proposed approach is able to parse the facial components in caricatures well. 

\vspace{-5mm}
\section{Conclusions}
\label{sec:majhead}
\vspace{-3mm}
In this work, we propose to transfer the knowledge from photo face parsing datasets to learn a caricature face parsing model by performing domain adaptation on the image level.
Specifically, we divide the adaptation into two stages in terms of shape and texture. 
We design a spatial transformer based network to achieve shape domain alignment and a style transfer network is followed to capture appearance-level domain shifts.
%
The synthesized images help train a caricature face parsing model that performs favorably against numerous baselines and state-of-the-art approaches.

\para{Acknowledgment.}
This work was supported in part by the National Nature Science Foundation of China (Grant Nos: 61751307) and in part by the National Youth Top-notch Talent Support Program.
W. Chu is sponsored by China Scholarship Council.

\bibliographystyle{IEEEbib}
\bibliography{egbib}
\end{document}